\def\year{2020}\relax
\documentclass[letterpaper]{article} 
\usepackage{aaai20}  
\usepackage{times}  
\usepackage{helvet} 
\usepackage{courier}  
\usepackage[hyphens]{url}  
\usepackage{graphicx} 
\usepackage{graphicx}  
\usepackage{latexsym}

\usepackage{array}
\usepackage{amssymb}
\usepackage{amsmath}
\usepackage{bm}
\usepackage{multirow}
\usepackage{subfig}
\usepackage{arydshln}
\usepackage{amsfonts}
\usepackage{graphicx}
\usepackage{graphics}
\usepackage{standalone}

\usepackage{bbm}
\usepackage{multicol}
\usepackage{multirow}

\usepackage{xspace}

\newcommand{\newcite}[1]{\citeauthor{#1}~(\citeyear{#1})}

\newcommand{\san}{\textsc{MlMhSan}s }

\title{Neuron Interaction Based Representation Composition \\ for Neural Machine Translation}

\author{
Jian Li,\textsuperscript{\rm 1,2}~
Xing Wang,\textsuperscript{\rm 3}~
Baosong Yang,\textsuperscript{\rm 4}~
Shuming Shi,\textsuperscript{\rm 3}~
Michael R. Lyu,\textsuperscript{\rm 1,2}~
Zhaopeng Tu\textsuperscript{\rm 3}\thanks{Corresponding author: Zhaopeng Tu. Work was partially done when Jian Li and Baosong Yang were interning at Tencent AI Lab.}\\
\textsuperscript{\rm 1}Department of Computer Science and Engineering, The Chinese University of Hong Kong\\
\textsuperscript{\rm 2}Shenzhen Research Institute, The Chinese University of Hong Kong\\
\{jianli,lyu\}@cse.cuhk.edu.hk\\
\qquad \qquad \textsuperscript{\rm 3}Tencent AI Lab \qquad \qquad \qquad 
\qquad \qquad \qquad \textsuperscript{\rm 4}University of Macau\\
\{brightxwang,shumingshi,zptu\}@tencent.com \qquad \qquad
nlp2ct.baosong@gmail.com
}

\begin{document}
\maketitle

\begin{abstract}

Recent NLP studies reveal that substantial linguistic information can be attributed to single neurons, i.e., individual dimensions of the representation vectors. 
We hypothesize that modeling strong interactions among neurons helps to better capture complex information by composing the linguistic properties embedded in individual neurons. 
Starting from this intuition, we propose a novel approach to compose representations learned by different components in neural machine translation
(e.g., multi-layer networks or multi-head attention), 
based on modeling strong interactions among neurons in the representation vectors.
Specifically, we leverage bilinear pooling to model pairwise multiplicative interactions among individual neurons, and a low-rank approximation to make the model computationally feasible.
We further propose extended bilinear pooling to incorporate first-order representations.
Experiments on WMT14 English$\Rightarrow$German and English$\Rightarrow$French translation tasks show that our model consistently improves performances over the SOTA \textsc{Transformer} baseline.
Further analyses demonstrate that our approach indeed captures more syntactic and semantic information as expected.
\end{abstract}

\section{Introduction}

Deep neural networks (\textsc{Dnn}s) have advanced the state of the art in various natural language processing (NLP) tasks, such as machine translation~\cite{Vaswani:2017:NIPS}, semantic role labeling~\cite{Strubell:2018:EMNLP}, and language representations~\cite{bert2018}. The strength of \textsc{Dnn}s lies in their ability to capture different linguistic properties of the input by different layers~\cite{Shi:2016:EMNLP,raganato2018analysis}, and composing (i.e. aggregating) these layer representations can further improve performances by providing more comprehensive linguistic information of the input~\cite{Peters:2018:NAACL,Dou:2018:EMNLP}.

Recent NLP studies show that single neurons in neural models which are defined as individual dimensions of the representation vectors, carry distinct linguistic information~\cite{Bau:2019:ICLR}. 
A follow-up work further reveals that simple properties such as coordinating conjunction (e.g., ``but/and'') or determiner (e.g., ``the'') can be attributed to individual neurons, while complex linguistic phenomena such as syntax (e.g., part-of-speech tag) and semantics (e.g., semantic entity type) are distributed across neurons~\cite{Dalvi:2019:AAAI}.
These observations are consistent with recent findings in neuroscience, which show that task-relevant information can be decoded from a group of neurons interacting with each other~\cite{Morcos:2016:Nature}. One question naturally arises: {\em can we better capture complex linguistic phenomena by composing/grouping the linguistic properties embedded in individual neurons?}

The starting point of our approach is an observation in neuroscience: {\em stronger neuron interactions} -- directly exchanging signals between neurons, enable more information processing in the nervous system~\cite{koch1983nonlinear}. We believe 
that simulating the neuron interactions in nervous system would be an appealing alternative to representation composition, which can potentially better learn the compositionality of natural language with subtle operations at a smaller granularity. 
Concretely, we employ bilinear pooling~\cite{Lin:2015:ICCV}, which executes pairwise multiplicative interactions among individual representation elements, to achieve \emph{strong} neuron interactions. 
We also introduce a low-rank approximation to make the original bilinear models computationally feasible~\cite{kim2016hadamard}.
Furthermore, as bilinear pooling only encodes multiplicative second-order features, we propose \emph{extended bilinear pooling} to incorporate first-order representations, which can capture more comprehensive information of the input sentences.

We validate the proposed neuron interaction based (NI-based) representation composition on top of multi-layer multi-head self-attention networks (\textsc{MlMhSan}s). The reason is two-fold. First, \textsc{MlMhSan}s are critical components of various SOTA \textsc{Dnn}s models, such as \textsc{Transformer}~\cite{Vaswani:2017:NIPS}, \textsc{Bert}~\cite{bert2018}, and \textsc{Lisa}~\cite{Strubell:2018:EMNLP}. Second, \textsc{MlMhSan}s involve in compositions
of both multi-layer representations and multi-head representations, which can investigate the universality of NI-based composition. Specifically, 
\begin{itemize}
    \item First, we conduct experiments on the machine translation task, a benchmark to evaluate the performance of neural models. Experimental results on the widely-used WMT14 English$\Rightarrow$German and English$\Rightarrow$French data show that the NI-based composition consistently improves performance over \textsc{Transformer} across language pairs. Compared with existing representation composition strategies~\cite{Peters:2018:NAACL,Dou:2018:EMNLP}, our approach shows its superiority in efficacy and efficiency.
    \item Second, we carry out linguistic analysis~\cite{conneau2018acl} on the learned representations from NMT encoder, and find that NI-based composition indeed captures more syntactic and semantic information as expected. These results provide support for our hypothesis that modeling strong neuron interactions helps to better capture complex linguistic information via advanced composition functions, which is essential for downstream NLP tasks.
\end{itemize}

This paper is an early step in exploring neuron interactions for representation composition in NLP tasks, which we hope will be a long and fruitful journey. 
We make the following contributions:
\begin{itemize}
    \item Our study demonstrates the necessity of modeling neuron interactions for representation composition in deep NLP tasks. We employ bilinear pooling to simulate the strong neuron interactions.
    \item We propose {\em extended bilinear pooling} to incorporate first-order representations, which produces a more comprehensive representation.
    \item Experimental results show that representation composition benefits the widely-employed \textsc{MlMhSan}s by aggregating information learned by multi-layer and/or multi-head attention components.
\end{itemize}

\section{Background}

\subsection{Multi-Layer Multi-Head Self-Attention}

In the past two years, \san based models establish the SOTA performances across different NLP tasks. The main strength of \san lies in the powerful representation learning capacity provided by the multi-layer and multi-head architectures. 
\san perform a series of nonlinear transformations from the input sequences to final output sequences. 

Specifically, \san are composed of a stack of $L$ identical layers ({\em multi-layer}), each of which is calculated as
\begin{eqnarray}
    {\bf H}^l &=& \textsc{Self-Att}({\bf H}^{l-1}) + {\bf H}^{l-1},
\label{eqn:enc}
\end{eqnarray}
where a residual connection is employed around each of two layers~\cite{he2016CVPR}.
$\textsc{Self-Att}(\cdot)$ is a self-attention model, which 
captures dependencies among hidden states in ${\bf H}^{l-1}$: 
\begin{eqnarray}
   \textsc{Self-Att}({\bf H}^{l-1}) = \textsc{Att}({\bf Q}^l, {\bf K}^{l-1}) \  {\bf V}^{l-1} \label{eq:out},
\end{eqnarray}
where $\{{\bf Q}^l, {\bf K}^{l-1}, {\bf V}^{l-1}\}$ 
are the query, key and value vectors that are transformed from the lower layer ${\bf H}^{l-1}$, respectively.

Instead of performing a single attention function,~\citeauthor{Vaswani:2017:NIPS}~\shortcite{Vaswani:2017:NIPS} found it is beneficial to capture different context features with multiple individual attention functions (\emph{multi-head}).
Concretely, multi-head attention model first transforms $\{{\bf Q}, {\bf K}, {\bf V}\}$ into $H$ subspaces with different, learnable linear projections:\footnote{Here we skip the layer index for simplification.}
\begin{equation}
  {\bf Q}_h, {\bf K}_h, {\bf V}_h   = {\bf Q}{\bf W}_h^{Q}, {\bf K}{\bf W}_h^{K}, {\bf V}{\bf W}_h^{V},  
\end{equation}
where $\{{\bf Q}_h, {\bf K}_h, {\bf V}_h\}$ are respectively the query, key, and value representations of the $h$-th head. $\{{\bf W}_h^{Q}, {\bf W}_h^{K}, {\bf W}_h^{V}\}$ 
denote parameter matrices associated with the $h$-th head. 
$H$ self-attention functions (Equation~\ref{eq:out}) are applied in parallel to produce the output states $\{{\bf O}_1,\dots, {\bf O}_H\}$. 
Finally, the $H$ outputs are concatenated and linearly transformed to produce a final representation:
\begin{eqnarray}
 \label{eq:concat_linear}
   {\bf H} = [{\bf O}_1, \dots, {\bf O}_H] \ {\bf W}^O, \label{eq:concat}
 \end{eqnarray}
where ${\bf W}^O \in \mathbb{R}^{d \times d}$ is a trainable matrix.

\subsection{Representation Composition}
Composing (i.e. aggregating) representations learned by different layers or attention heads has been shown beneficial for \san\cite{Dou:2018:EMNLP,Ahmed:2018:arXiv}.
Without loss of generality, from here on, we refer to $\{{\bf r}_1, \dots, {\bf r}_N\} \in \mathbb{R}^{d}$ for the representations to compose, where ${\bf r}_i$ can be a layer representation (${\bf H}^l$, Equation~\ref{eqn:enc}) or head representation (${\bf O}_h$, Equation~\ref{eq:concat}). The composition is expressed as
\begin{eqnarray}
    \mathbf{\widetilde{H}} = \textsc{Compose}({\bf r}_1, \dots, {\bf r}_N),
    \label{eqn:neural}
\end{eqnarray}
where $\textsc{Compose}(\cdot)$ can be arbitrary functions, such as linear combination\footnote{The linear composition of multi-head representations (Equation~\ref{eq:concat}) can be rewritten in the format of weighted sum: ${\bf O}=\sum_{h=1}^H {\bf O}_h {\bf W}^O_h$ with ${\bf W}^O_h \in \mathbb{R}^{\frac{d}{H} \times d}$.}~\cite{Peters:2018:NAACL,Ahmed:2018:arXiv} and hierarchical aggregation~\cite{Dou:2018:EMNLP}. 
Although effective to some extent, these approaches do not model neuron interactions among the representation vectors, which we believe is valuable for representation composition in deep NLP models.

\section{Approach}

\begin{figure*}[t]
\centering
\subfloat[Bilinear Pooling]{
 \includegraphics[width=0.42\textwidth]{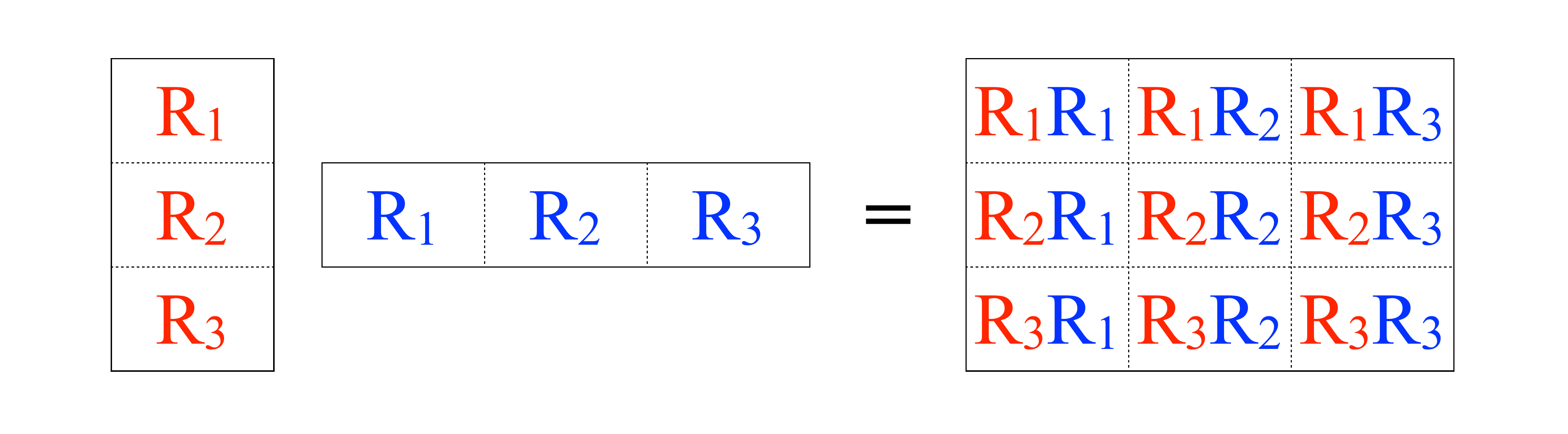}
} \hspace{0.1\textwidth}
\subfloat[Extended Bilinear Pooling]{
 \includegraphics[width=0.42\textwidth]{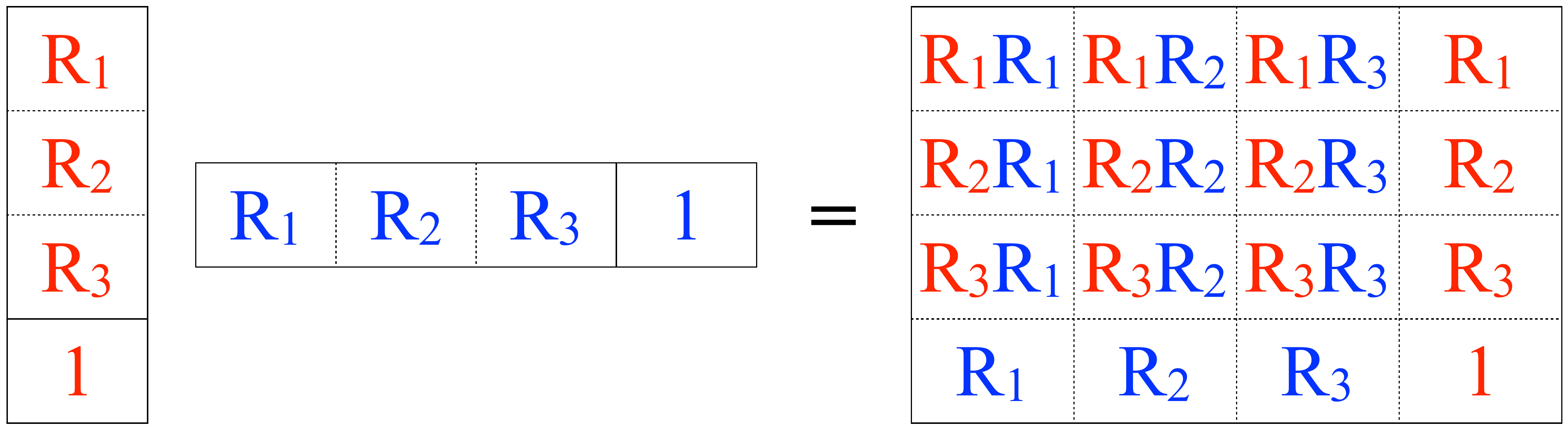}
}
\caption{Illustration of (a) {\em bilinear pooling} that models fully neuron-wise multiplicative interaction, and (b) {\em extended bilinear pooling} that captures both second- and first-order neuron interactions.}
\label{fig:bilinear-pooling}
\end{figure*}

\subsection{Motivation}

Different types of neurons in the nervous system carry distinct signals~\cite{cohen2012nature}. Similarly, neurons in deep NLP models -- individual dimensions of representation vectors, carry distinct linguistic information~\cite{Bau:2019:ICLR,Dalvi:2019:AAAI}.
Studies in neuroscience reveal that stronger neuron interactions bring more information processing capability~\cite{koch1983nonlinear},
which we believe also applies to deep NLP models.

In this work, we explore the strong neuron interactions provided by bilinear pooling for representation composition. Bilinear pooling~\cite{Lin:2015:ICCV} is a recently proposed feature fusion approach in the vision field.
Instead of linearly combining all representations, bilinear pooling executes pairwise multiplicative interactions among individual representations, to model \emph{full} neuron interactions as shown in Figure~\ref{fig:bilinear-pooling}(a).

Note that there are many possible ways to implement the neuron interactions.
The aim of this paper is not to explore this whole space but simply to show that one fairly straightforward implementation works well on a strong benchmark.

\subsection{Bilinear Pooling for Neuron Interaction}
\label{sec:LRBP}

\paragraph{Bilinear Pooling}
Bilinear pooling~\cite{Tenenbaum:2000:NeuroComputation} is defined as an \emph{outer product} of two representation vectors followed by a linear projection. As illustrated in Figure~\ref{fig:bilinear-pooling}(a), all elements of the two vectors have direct multiplicative interactions with each other. However, in the scenario of multi-layer and multi-head composition, we generally have more than two representation vectors to compose (i.e., $L$ layers and $H$ attention heads). To utilize the full second-order (i.e. multiplicative) interactions in bilinear pooling, we concatenate all the representation vectors and feed the concatenated vector twice to the bilinear pooling.
Concretely, we have:
\begin{eqnarray}
    {\bf R} &=& | \widehat{\bf R} \widehat{\bf R}^{\top} | {\bf W}^B, \\
    \widehat{\bf R} &=& [{\bf r}_1, \dots, {\bf r}_N],
\label{eq:con}
\end{eqnarray}
where $|\widehat{\bf R} \widehat{\bf R}^{\top}| \in \mathbb{R}^{Nd \times Nd}$ is the outer product of the concatenated representation $\widehat{\bf R}$, $|\cdot|$ denotes serializing the matrix into a vector with dimensionality $(Nd)^2$. In this way, all elements in the partial representations are able to interact with each other in a multiplicative way. 

However, the parameter matrix ${\bf W}^B \in \mathbb{R}^{(Nd)^2 \times d}$ and computing cost cubically increases with dimensionality $d$, which becomes problematic when training or decoding on a GPU with limited memory\footnote{For example, a regular \textsc{Transformer} model requires a huge amount of 36 billion ($(Nd)^2 \times d$) parameters for $d=1000$ and $N=6$.}. 
There have been a few attempts to reduce the computational complexity of the original bilinear pooling.~\newcite{Gao:2016:CVPR} propose {\em compact bilinear pooling} to reduce the quadratic expansion of dimensionality for image classification.~\newcite{kim2016hadamard} and~\newcite{Kong:2017:CVPR} propose {\em low-rank bilinear pooling} for visual question answering and image classification respectively, which further reduces the parameters to be learned and achieves comparable effectiveness with full bilinear pooling. In this work, we focus on the low-rank approximation for its efficiency, and generalize from the original model for deep representations.


\paragraph{Low-Rank Approximation}
In the full bilinear models, each output element $R_i \in \mathbb{R}^{1}$ can be expressed as
\begin{eqnarray}
R_i &=& \sum_{j=1}^{Nd} \sum_{k=1}^{Nd} { w}^B_{jk,i} \widehat{ R}_j \widehat{ R}^{\top}_k  \nonumber \\
          &=& \widehat{\bf R}^{\top} {\bf W}^B_i \widehat{\bf R}    \label{eq:full-bilinear},
\end{eqnarray}
where ${\bf W}^B_i \in \mathbb{R}^{Nd \times Nd}$ is a weight matrix to produce output element $ R_i$.
The low-rank approximation enforces the rank of ${\bf W}^B_i$ to be low-rank $r \leq Nd$ ~\cite{pirsiavash2009bilinear}, which is then factorized as ${\bf U}_i {\bf V}_i^{\top}$ with ${\bf U}_i \in \mathbb{R}^{Nd \times r}$ and ${\bf V}_i \in \mathbb{R}^{Nd \times r}$. Accordingly, Equation~\ref{eq:full-bilinear} can be rewritten as
\begin{eqnarray}
    {R}_i &=& \widehat{\bf R}^{\top} {\bf U}_i {\bf V}_i^{\top} \widehat{\bf R} \nonumber \\
    &=& (\widehat{\bf R}^{\top} {\bf U}_i \odot \widehat{\bf R}^{\top}{\bf V}_i) \mathbbm{1}_r,
    \label{eq:low-rank}
\end{eqnarray}
where $\mathbbm{1}_r$ is a $r$-dimensional vector of ones, $\odot$ represents element-wise product. 
By replacing $\mathbbm{1}_r$ with ${\bf P} \in \mathbb{R}^{r\times d}$, and redefining ${\bf U} \in \mathbb{R}^{Nd\times r}$ and ${\bf V} \in \mathbb{R}^{Nd \times r}$, the low-rank approximation can be defined as
\begin{equation}
    {\bf R} = (\widehat{\bf R}^{\top}{\bf U} \odot \widehat{\bf R}^{\top}{\bf V}) {\bf P}. 
\label{eq:final}
\end{equation}

In this way, the computation complexity is reduced from $O(d^3)$ to $O(d^2)$. And the parameter matrices {\bf U}, {\bf V}, and {\bf P} are now feasible to fit in GPU memory.

\paragraph{Extended Bilinear Pooling with First-Order Representation}
Previous work in information theory has proven that second-order and first-order representations encode different types of information~\cite{goudreau1994firstorder}, which we believe also holds on NLP tasks.
As bilinear pooling only encodes second-order (i.e., multiplicative) interactions among individual neurons, we propose the \emph{extended bilinear pooling} to inherit the advantages of first-order representations and form a more comprehensive representation.

Specifically, we append $\mathbf{1}$s to the representation vectors. As illustrated in Figure~\ref{fig:bilinear-pooling}(b), we respectively append $\mathbf{1}$ to the two ${\bf R}$ vectors, then the outer product of them produces both second-order and first-order interactions among the elements. According to Equation~\ref{eq:final}, the final representation is revised as: 
\begin{equation}
    {\bf R_f} = (\begin{bmatrix}\widehat{\bf R}\\1\end{bmatrix}^{\top} {\bf U} \odot \begin{bmatrix}\widehat{\bf R}\\1\end{bmatrix}^{\top} {\bf V})~ {\bf P},
\end{equation}
where $\widehat{\bf R}$ is the concatenated representation as in Equation~\ref{eq:con}.
As a result, the final representation $\bf R_f$ preserves both multiplicative bilinear features (as in Equation~\ref{eq:final}) and first-order linear features (as in Equation~\ref{eq:concat_linear}).

\begin{table*}[t]
  \centering
  \begin{tabular}{c|l||r c c||c}
    {\bf \#}    &   {\bf Model} &  \bf {\# Para.} & \bf {Train}  &   \bf Decode  &    \bf  BLEU\\
    \hline
    1   &   \textsc{Transformer-Base}	  & 88.0M	&   2.02  & 1.50 &  $27.31$\\   
    \hline
    \multicolumn{6}{c}{\em Existing representation composition} \\
    \hline
    2   & ~~+ Multi-Layer: Linear Combination  & +3.1M &1.98 &1.46 & $27.77$   \\
    \hdashline
    3   &~~+ Multi-Layer: Hierarchical Aggregation &  +23.1M   &   1.62    &   1.36  &  $28.32$\footnotemark \\
    4   &~~+ Multi-Head: Hierarchical Aggregation  & +13.6M & 1.74   & 1.38 &  28.13   \\
    5   &~~+ Both (3+4) & +36.7M & 1.42 & 1.25 & 28.42 \\
    \hline
    \multicolumn{6}{c}{\em This work: neuron-interaction based representation composition} \\
    \hline
    6   &   ~~+ Multi-Layer: {\em NI-based Composition}   & +16.8M & 1.93 & 1.44& $28.31$\\
    7   &   ~~+ Multi-Head: {\em NI-based Composition}    & +14.1M & 1.92  &1.43 &    $28.29$\\
    8   &   ~~+ Both (6+7) &+30.9M & 1.87& 1.40 & {\bf 28.54}   \\
  \end{tabular}
  \caption{Translation performance on WMT14 English$\Rightarrow$German
  translation task. ``\# Para.'' denotes the number of parameters, and ``Train'' and ``Decode'' respectively denote the training speed (steps/second) and decoding speed (sentences/second). We compare our model with linear combination~\cite{Peters:2018:NAACL} and hierarchical aggregation~\cite{Dou:2018:EMNLP}. } 
  \label{tab:comparison}
\end{table*}

\paragraph{Applying to \textsc{Transformer}}
\textsc{Transformer}~\cite{Vaswani:2017:NIPS} consists of an encoder and a decoder, each of which is stacked in 6 layers where we can apply multi-layer composition (excluding the embedding layer) to produce the final representations of the encoder and decoder. Besides, each layer has one (in encoder) or two (in decoder) multi-head attention component with $H$ heads, to which we can apply multi-head composition to substitute Equation~\ref{eq:concat}. The two sorts of representation composition can be used individually, while combining them is expected to further improve the performance.
\footnotetext{The original result in \cite{Dou:2018:EMNLP} is $28.63$, which is \emph{case-insensitive}. As we report case-sensitive BLEU scores, we have requested \citeauthor{Dou:2018:EMNLP} to get this result.}

\section{Experiments}

\begin{table*}[t]
  \centering
  \begin{tabular}{l||rcl|rcl}
    \multirow{2}{*}{\bf Architecture}  &  \multicolumn{3}{c|}{\bf EN$\Rightarrow$DE}  & \multicolumn{3}{c}{\bf EN$\Rightarrow$FR}\\
    \cline{2-7}
        &   \# Para.    &   Train    &   BLEU    &   \# Para.   &   Train     &   BLEU\\
    \hline \hline
    \multicolumn{7}{c}{{\em Existing NMT systems}: {\cite{Vaswani:2017:NIPS}}} \\
    \hline
    \textsc{Transformer-Base}    &  65M  &   n/a  &   $27.3$  &   n/a   &   n/a&   $38.1$ \\ 
    \textsc{Transformer-Big}    &  213M  &  n/a &  $28.4$ &   n/a  &   n/a  &   $41.8$\\ 
    \hline\hline
    \multicolumn{7}{c}{{\em Our NMT systems}}   \\ \hline
    \textsc{Transformer-Base}  &  88M  &   2.02  &   $27.31$   & 95M & 2.01& $39.28$  \\
    ~~~ + NI-Based Composition   & 118M & 1.87&  $28.54^\Uparrow$ &125M  &1.85 & $40.15^\Uparrow$  \\
    \hline
    \textsc{Transformer-Big}    & 264M  &  0.85 &   $28.58$ &278M &0.84 & $41.41$  \\
    ~~~ + NI-Based Composition   & 387M & 0.61 &  $29.17^\Uparrow$ & 401M & 0.59 & $42.10^\Uparrow$\\
  \end{tabular}
  \caption{Comparing with existing NMT systems on WMT14 English$\Rightarrow$German (``EN$\Rightarrow$DE'') and English$\Rightarrow$French (``EN$\Rightarrow$FR'') translation tasks.
  ``$\Uparrow$'': significantly better than the baseline ($p < 0.01$) using bootstrap resampling~\cite{Koehn2004Statistical}.}
  \label{tab:main}
\end{table*}

\subsection{Setup}

\paragraph{Dataset}
We conduct experiments on the WMT2014 English$\Rightarrow$German (En$\Rightarrow$De) and English$\Rightarrow$French (En$\Rightarrow$Fr) translation tasks. The En$\Rightarrow$De dataset consists of about 4.56 million sentence pairs. We use newstest2013 as the development set and newstest2014 as the test set. 
The En$\Rightarrow$Fr dataset consists of $35.52$ million sentence pairs. We use the concatenation of newstest2012 and newstest2013 as the development set and newstest2014 as the test set.
We employ BPE~\cite{sennrich2016neural} with 32K merge operations for both language pairs.
We adopt the case-sensitive 4-gram NIST BLEU score~\cite{papineni2002bleu} as our evaluation metric and bootstrap resampling~\cite{Koehn2004Statistical} for statistical significance test.

\paragraph{Models}
We evaluate the proposed approaches on the advanced \textsc{Transformer} model~\cite{Vaswani:2017:NIPS}, and implement on top of an open-source toolkit -- THUMT~\cite{zhang2017thumt}. We follow~\citeauthor{Vaswani:2017:NIPS}~\shortcite{Vaswani:2017:NIPS} to set the configurations and have reproduced their reported results on the En$\Rightarrow$De task. The parameters of the proposed models are initialized by the pre-trained \textsc{Transformer} model.
We have tested both \emph{Base} and \emph{Big} models, which differ at hidden size (512 vs. 1024) and number of attention heads (8 vs. 16). Concerning the low-rank parameter (Equation~\ref{eq:low-rank}), we set low-rank dimensionality $r$ to 512 and 1024 in \emph{Base} and \emph{Big} models respectively. All models are trained on eight NVIDIA P40 GPUs where each is allocated with a batch size of 4096 tokens. In consideration of computation cost, we study model variations with \emph{Base} model on the En$\Rightarrow$De task, and evaluate overall performance with \emph{Big} model on both En$\Rightarrow$De and En$\Rightarrow$Fr tasks.

\subsection{Comparison to Existing Approaches}
In this section, we evaluate the impacts of different representation composition strategies on the En$\Rightarrow$De translation task with \textsc{Transformer-Base}, as listed in Table~\ref{tab:comparison}.

\paragraph{Existing Representation Composition} (Rows 1-5)
For the conventional \textsc{Transformer} model, it adopts multi-head composition with linear combination but only uses top-layer representation as its default setting.
Accordingly, we keep the linear multi-head composition (Row 1) unchanged, and choose two representative multi-layer composition strategies (Rows 2 and 3):  the widely-used linear combination~\cite{Peters:2018:NAACL} and the effective hierarchical aggregation~\cite{Dou:2018:EMNLP}. The hierarchical aggregation merges states of different layers through a CNN-like tree structure with the filter size being two, to hierarchically preserve and combine feature channels.

As seen, linearly combining all layers (Row 2) achieves +0.46 BLEU improvement over \textsc{Transformer-Base} with almost the same training and decoding speeds.
Hierarchical aggregation for multi-layer composition (Row 3)  yields larger improvement in terms of BLEU score, but at the cost of considerable speed decrease.
To make a fair comparison, we also implement hierarchical aggregation for multi-head composition (Rows 4 and 5), which consistently improves performances at the cost of introducing more parameters and slower speeds.

\paragraph{The Proposed Approach} (Rows 6-8)
Firstly, we apply our NI-based composition, i.e. \emph{extended bilinear pooling},  for multi-layer composition with the default linear multi-head composition (Row 6). We find that the approach  achieves almost the same translation performance as hierarchical aggregation (Row 3), while keeps the training and decoding speeds as \emph{efficient} as linear combination. 
Then, we apply the NI-based approach for multi-head composition with the default top layer exploitation (Row 7).  We can see that our approach gains  +0.98 BLEU point over \textsc{Transformer-Base} and achieves more improvement than hierarchical aggregation (Row 4). The two results demonstrate that our NI-based approach can be effectively applied to different representation composition scenarios.

At last, we simultaneously apply the NI-based approach to the multi-layer and multi-head composition (Row 8). Our model achieves further improvement over individual models and the hierarchical aggregation (Row 5), showing that \textsc{Transformer} can benefit from the complementary composition from multiple heads and historical layers. In the following experiments, we adopt NI-based composition for both the multi-layer and multi-head compositions as the default strategy.

\subsection{Main Results on Machine Translation}

In this section, we validate the proposed NI-based representation composition on both WMT14 En$\Rightarrow$De and En$\Rightarrow$Fr translation tasks. Experimental results are listed in Table~\ref{tab:main}. The performances of our implemented \textsc{Transformer} match the results on both language pairs reported in previous work~\cite{Vaswani:2017:NIPS}, which we believe makes the evaluation convincing. 

Incorporating NI-based composition consistently and significantly improves translation performance for both base and big \textsc{Transformer} models across language pairs, demonstrating the effectiveness and universality of the proposed NI-based representation composition. It is encouraging to see that  \textsc{Transformer-Base} with  NI-based composition even achieves competitive performance as that of \textsc{Transformer-Big} in the En$\Rightarrow$De task, with only half fewer parameters and the training speed is twice faster. This further demonstrates that our performance gains are not simply brought by additional parameters. Note that the improvement on En$\Rightarrow$De task is larger than En$\Rightarrow$Fr task, which can be attributed to the size of training data (4M vs. 35M). 

\begin{table}[t]
\centering
\begin{tabular}{c|c||c | c | c}
  \multicolumn{2}{c||}{\bf Task} & {\bf Base} &	{\bf \textsc{Ours}} &  \bf $\bigtriangleup$\\
  \hline\hline
  \multirow{3}{*}{\rotatebox[origin=c]{90}{{\bf Surface}}}       
  &	SeLen	&  92.20 &  92.11   &   -0.1\%\\
  &	WC		&  63.00 &  63.50   &   +0.8\%\\
  \cdashline{2-5}
  &	Ave.    & 77.60  &  77.81 & +0.3\%\\
  \hline
  \multirow{4}{*}{\rotatebox[origin=c]{90}{{\bf Syntactic}}}    
  &	TrDep	&  44.74 & 44.96 &  +0.5\%\\
  &	ToCo	&  79.02 & 81.31 & \bf +2.9\%\\
  &	BShif   &  71.24 & 72.44 & \bf +1.7\%\\
  \cdashline{2-5}
  &	Ave.    &  65.00 & 66.24 & \bf +1.9\%\\
  \hline
  \multirow{6}{*}{\rotatebox[origin=c]{90}{{\bf Semantic}}}    
  &	Tense	&  89.24 &  89.26   &   +0.0\%\\
  &	SubNm	&  84.69 &  87.05   & \bf +2.8\%\\
  &	ObjNm	&  84.53 &  86.91   & \bf +2.8\%\\
  &	SOMO	&  52.13 &  52.52   & +0.7\%\\
  &	CoIn	&  62.47 &  64.93   & \bf +3.9\%\\
  \cdashline{2-5}
  &	Ave.    &  74.61 &  76.13   & \bf +2.0\%\\ 
\end{tabular}
  \caption{Classification accuracies on 10 probing tasks of evaluating the linguistic properties (``Surface'', ``Syntactic'', and ``Semantic''). ``Ave.'' denotes the averaged accuracy in each category. ``$\bigtriangleup$'' denotes the relative improvement, and we highlight the numbers $\geq 1\%$.}
\label{tab:probing}
\end{table}

\subsection{Analysis}
In this section, we conduct extensive analysis to deeply understand the proposed models in terms of 1) investigating the linguistic properties learned by the NMT encoder; 2) the influences of first-order representation and low-rank constraint; and 3) the translation performances on sentences of varying lengths.

\paragraph{Targeted Linguistic Evaluation on NMT Encoder}
Machine translation is a complex task, which consists of both the understanding of input sentence (encoder) and the generation of output conditioned on such understanding (decoder). In this probing experiment, we evaluate the understanding part using Transformer encoders that are trained on the EN$\Rightarrow$DE NMT data, and are fixed in the probing tasks with only MLP classifiers being trained on probing data. 

Recently,~\newcite{conneau2018acl} designed 10 probing tasks to study what linguistic properties are captured by representations from sentence encoders.
A probing task is a classification problem that focuses on simple linguistic properties of input sentences, including surface information, syntactic information, and semantic information.
For example, ``WC'' tests whether it is possible to recover information about the original words given its sentence embedding. ``Bshif'' checks whether two consecutive tokens have been inverted. ``SubNm'' focuses on the number of the subject of the main clause.
For more detailed description about the 10 tasks, interested readers can refer to the original paper~\cite{conneau2018acl}.  
We conduct probing tasks to examine whether the NI-based representation composition can benefit the \textsc{Transformer} encoder to produce more informative representation.

Table~\ref{tab:probing} lists the results.
The NI-based composition outperforms that by the baseline in most probing tasks, proving that our composition strategy indeed helps \textsc{Transformer} encoder generate more informative representation, especially at the syntactic and semantic level. The averaged gains in syntactic and semantic tasks are significant, showing that our strategy makes \textsc{San} capture more high-level linguistic properties. Note that the lower values in surface tasks (e.g., SeLen), are consistent with the conclusion in \cite{conneau2018acl}: as model captures deeper linguistic properties, it will tend to forget about these superficial features.

\begin{figure}[h]
\centering
\includegraphics[width=0.35\textwidth]{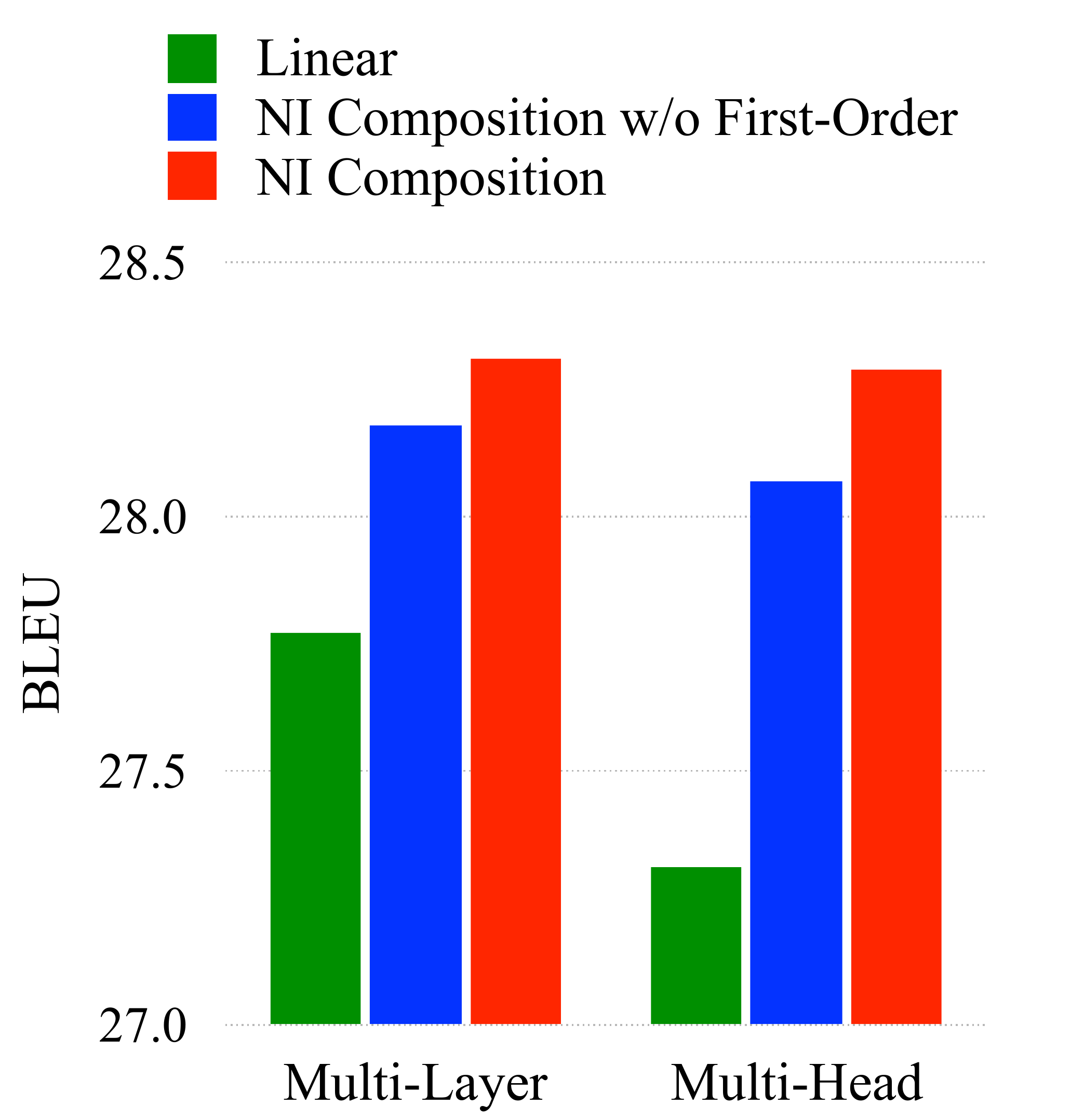}
\caption{Effect of first-order representation on WMT14 En$\Rightarrow$De translation task.}
\label{fig:residual}
\end{figure}

\paragraph{Effect of First-Order Representation}
As aforementioned, we extend the conventional bilinear pooling by appending $\mathbf{1}$s to the representation vectors thus incorporate first-order representations (i.e. linear combination), and capture both multiplicative bilinear features and additive linear features. Here we conduct ablation study to validate the effectiveness of each component. We respectively experiment on multi-layer and multi-head representation composition, and the results are shown in Figure~\ref{fig:residual}. 

Several observations can be made.
First, we notice that by replacing linear combination with mere bilinear pooling (``NI-based composition w/o first-order'' in Figure~\ref{fig:residual}), the translation performance significantly improves both in multi-layer and multi-head composition, demonstrating the effectiveness of full neuron interaction and second-order features.
We further observe that it is indeed beneficial to extend bilinear pooling with linear combination (``NI composition'' in Figure~\ref{fig:residual}) which captures the complementary information among them and forms a more comprehensive representation of the input.

\begin{figure}[h]
  \centering
\includegraphics[width=0.4\textwidth]{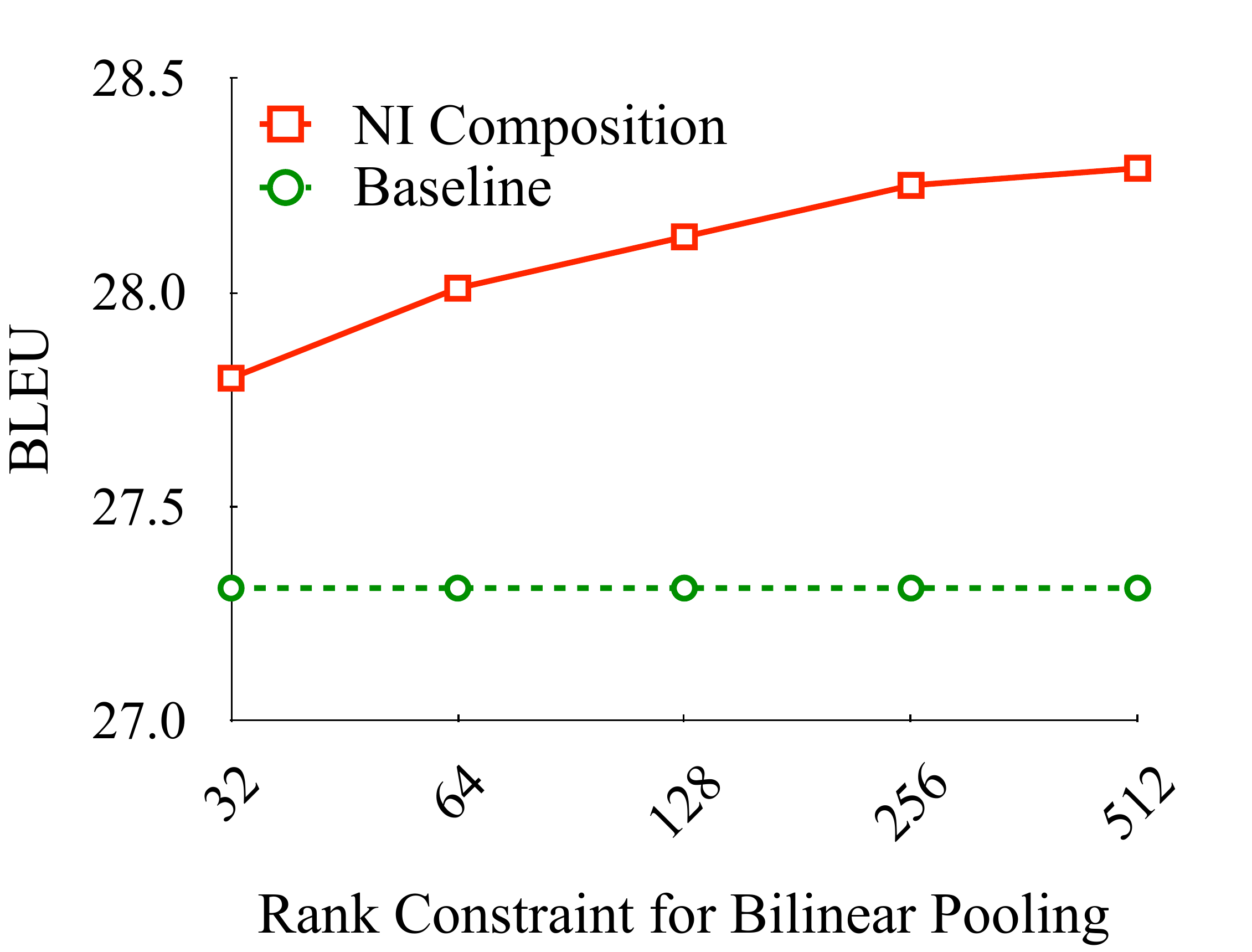}
\caption{BLEU scores on the En$\Rightarrow$De test set with different rank constraints for bilinear pooling. ``Baseline'' denotes \textsc{Transformer-Base}.}
\label{fig:low-rank}
\end{figure}

\paragraph{Effect of Low-Rank Constraint} In this experiment, we study the impact of low-rank constraint $r$ (Equation~\ref{eq:low-rank}) on bilinear pooling, as shown in Figure~\ref{fig:low-rank}.
It is interesting to investigate whether the model with a smaller setting of $r$ can also achieve considerable results.
We examine groups of multi-head composition models with different $r$ on the En$\Rightarrow$De translation task. From Figure~\ref{fig:low-rank}, we can see that the translation performance increases with larger $r$ value and the model with $r=51$2 achieves best performance\footnote{The maximum value of $r$ is 512 since the rank of a matrix ${\bf W} \in \mathbb{R}^{Nd \times Nd}$ is bounded by $Nd$.}. Note that even when the dimensionality $r$ is reduced to 32, our model can still consistently outperform the baseline with only 0.9M parameters added (not shown in the figure). This reconfirms our claim that the improvements on the BLEU score could not be simply attributed to the additional parameters.

\begin{figure}[h]
\centering
\includegraphics[width=0.4\textwidth]{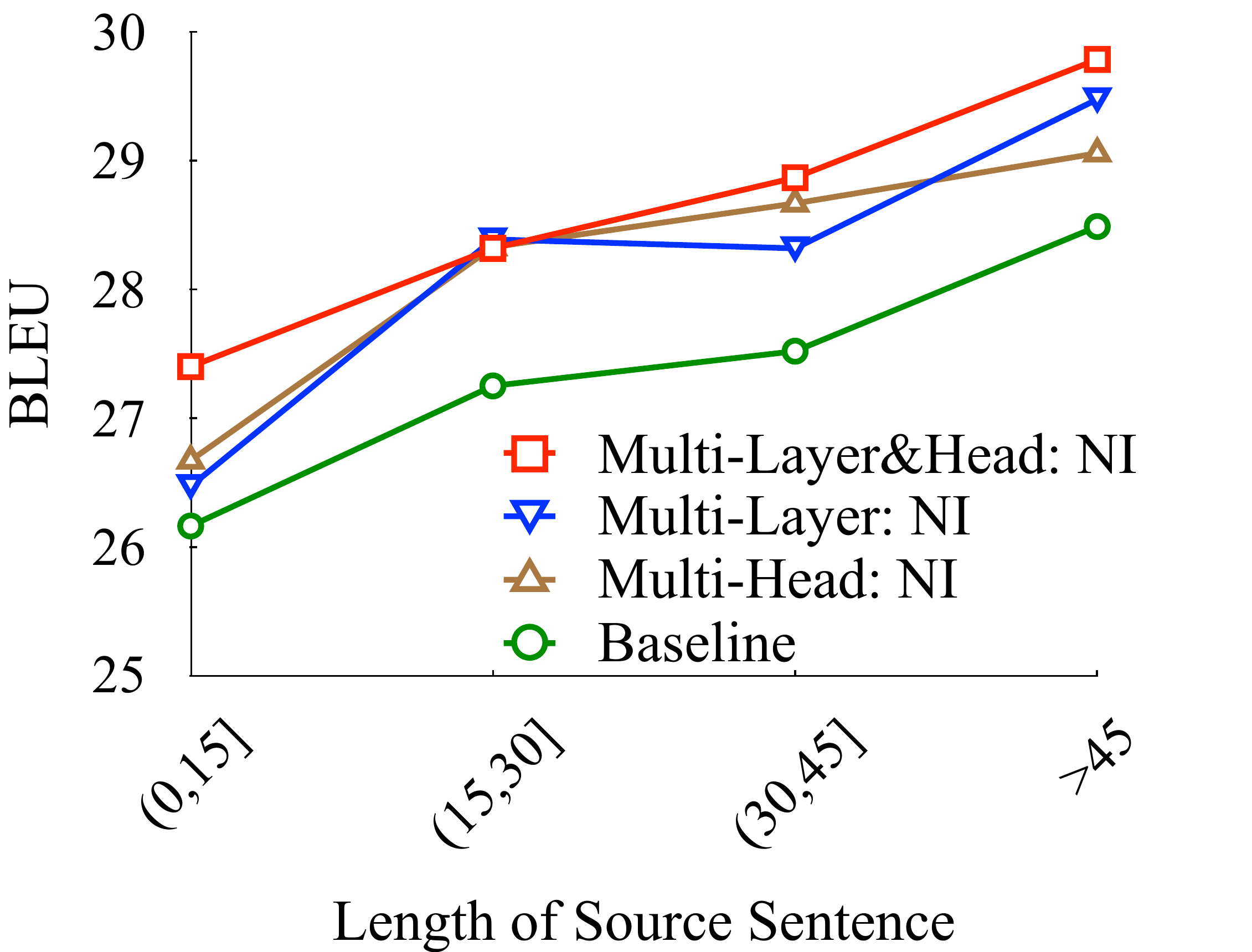}
\caption{BLEU scores on the En$\Rightarrow$De test set with respect to various input sentence lengths. ``Baseline'' denotes \textsc{Transformer-Base}.}
\label{fig:length}
\end{figure}

\paragraph{Length Analysis}
We group sentences of similar lengths together and compute the BLEU score for each group, as shown in Figure~\ref{fig:length}. Generally, the performance of \textsc{Transformer} goes up with the increase of input sentence lengths, which is different from the results on single-layer RNNSearch models (i.e., performance decreases on longer sentences) as shown in~\cite{tu2016modeling}. We attribute this phenomenon to the advanced \textsc{Transformer} architecture including multiple layers, multi-head attention and feed-forward networks.

Clearly, our NI-based approaches outperform the baseline \textsc{Transformer} in all length segments, including only using multi-layer composition or multi-head composition, which verifies our contribution that representation composition indeed benefits \textsc{San}s. Moreover, multi-layer composition and multi-head composition are complementary to each other regarding different length segments, and simultaneously applying them achieves further performance gain. 

\section{Related Work}


\paragraph{Bilinear Pooling}
Bilinear pooling has been well-studied in the computer vision community,  which is first introduced by \newcite{Tenenbaum:2000:NeuroComputation} to separate style and content. Bilinear pooling has since then been considered to replace fully-connected layers in neural networks by introducing second-order statistics, and applied to fine grained recognition~\cite{Lin:2015:ICCV}. While bilinear models provide richer representations than linear models~\cite{goudreau1994firstorder}, bilinear pooling produces a high-dimensional feature of quadratic expansion, which may constrain model structures and computational resources. To address this challenge, \newcite{Gao:2016:CVPR} propose compact bilinear pooling through random projections for image classification, which is further applied to visual question answering~\cite{fukui2016multimodal}. 
~\newcite{kim2016hadamard} and~\newcite{Kong:2017:CVPR} independently propose low-rank approximation on the transformation matrix of bilinear pooling, which aims to reduce the model size and corresponding computational burden. Their models are applied to visual question answering and fine-grained image classification, respectively.

While most work focus on computer vision tasks, our work is among the few studies~\cite{dozat2017ICLR,Delbrouck:2017:arXiv}, which prove the idea of bilinear pooling can have promising applications on NLP tasks. Our approach differs at: 1) we apply bilinear pooling to representation composition in NMT, while they apply to the attention model in either parsing or multimodal NMT; and 2) we extend the original bilinear pooling to incorporate first-order representations, which consistently improves translation performance in different scenarios (Figure~\ref{fig:residual}).

\paragraph{Multi-Layer Representation Composition}


Exploiting multi-layer representations has been well studied in the NLP community. \newcite{Peters:2018:NAACL} have found that linearly combining different layers is helpful and improves their performances on various NLP tasks.
In the context of NMT, several neural network based approaches to fuse information across historical layers have been proposed, such as dense information flow~\cite{Shen:2018:NAACL}, iterative and hierarchical aggregation~\cite{Dou:2018:EMNLP}, routing-by-agreement~\cite{Dou:2019:AAAI}, and transparent attention~\cite{Bapna:2018:EMNLP}.

In this work, we consider representation composition from a novel perspective of \emph{modeling neuron interactions}, which we prove is a promising and effective direction.
Besides, we generalize layer aggregation to representation composition in \textsc{San}s by also considering multi-head composition, and we propose an unified NI-based approach to aggregate both types of representation.


\paragraph{Multi-Head Self-Attention}

Multi-head attention has shown promising results in many NLP tasks, such as machine translation~\cite{Vaswani:2017:NIPS} and semantic role labeling~\cite{Strubell:2018:EMNLP}. The strength of multi-head attention lies in the rich expressiveness by using multiple attention functions in different representation subspaces. 
Previous work show that multi-head attention can be further enhanced by encouraging individual attention heads to extract distinct information. 
For example,~\newcite{Li:2018:EMNLP} propose disagreement regularizations to encourage different attention heads to encode distinct features, and~\newcite{Strubell:2018:EMNLP} employ different attention heads to capture different linguistic features.
\newcite{Li:2019:NAACL} is a pioneering work on empirically validating the importance of information aggregation for multi-head attention. Along the same direction, we apply the NI-based approach to compose the representations learned by different attention heads (as well as different layers), and empirically reconfirm their findings.


\section{Conclusion}

In this work, we propose NI-based representation composition for \textsc{MlMhSan}s, by modeling strong neuron interactions in the representation vectors generated by different layers and attention heads. Specifically, we employ bilinear pooling to capture pairwise multiplicative interactions among individual neurons, and propose \emph{extended bilinear pooling} to further incorporate first-order representations.
Experiments on machine translation tasks show that our approach effectively and efficiently improves translation performance over the \textsc{Transformer} model, and multi-head composition and multi-layer composition are complementary to each other. Further analyses reveal that our model makes the encoder of \textsc{Transformer} capture more syntactic and semantic properties of input sentences.

Future work includes exploring more neuron interaction based approaches for representation composition other than the bilinear pooling, and applying our model to a variety of network architectures such as \textsc{Bert}~\cite{bert2018} and \textsc{Lisa}~\cite{Strubell:2018:EMNLP}.

\section{Acknowledgement}
The work described in this paper was partially supported by the Research Grants Council of the Hong Kong Special Administrative Region, China (No. CUHK 14210717 of the General Research Fund), and Microsoft Research Asia (2018 Microsoft Research Asia Collaborative Research Award). We thank the anonymous reviewers for their comments and suggestions.

\bibliography{all}

\begin{thebibliography}{}

\bibitem[\protect\citeauthoryear{Ahmed, Keskar, and
  Socher}{2018}]{Ahmed:2018:arXiv}
Ahmed, K.; Keskar, N.~S.; and Socher, R.
\newblock 2018.
\newblock {Weighted Transformer Network for Machine Translation}.
\newblock {\em arXiv}.

\bibitem[\protect\citeauthoryear{Bapna \bgroup et al\mbox.\egroup
  }{2018}]{Bapna:2018:EMNLP}
Bapna, A.; Chen, M.; Firat, O.; Cao, Y.; and Wu, Y.
\newblock 2018.
\newblock Training deeper neural machine translation models with transparent
  attention.
\newblock In {\em EMNLP}.

\bibitem[\protect\citeauthoryear{Bau \bgroup et al\mbox.\egroup
  }{2019}]{Bau:2019:ICLR}
Bau, A.; Belinkov, Y.; Sajjad, H.; Durrani, N.; Dalvi, F.; and Glass, J.
\newblock 2019.
\newblock Identifying and controlling important neurons in neural machine
  translation.
\newblock In {\em ICLR}.

\bibitem[\protect\citeauthoryear{Cohen \bgroup et al\mbox.\egroup
  }{2012}]{cohen2012nature}
Cohen, J.~Y.; Haesler, S.; Vong, L.; Lowell, B.~B.; and Uchida, N.
\newblock 2012.
\newblock Neuron-type-specific signals for reward and punishment in the ventral
  tegmental area.
\newblock {\em Nature} 482(7383):85.

\bibitem[\protect\citeauthoryear{Conneau \bgroup et al\mbox.\egroup
  }{2018}]{conneau2018acl}
Conneau, A.; Kruszewski, G.; Lample, G.; Barrault, L.; and Baroni, M.
\newblock 2018.
\newblock {What You Can Cram into A Single ${\$}{\&}!{\#}*$ Vector: Probing
  Sentence Embeddings for Linguistic Properties}.
\newblock In {\em ACL}.

\bibitem[\protect\citeauthoryear{Dalvi \bgroup et al\mbox.\egroup
  }{2019}]{Dalvi:2019:AAAI}
Dalvi, F.; Durrani, N.; Sajjad, H.; Belinkov, Y.; Bau, A.; and Glass, J.
\newblock 2019.
\newblock What is one grain of sand in the desert? analyzing individual neurons
  in deep nlp models.
\newblock In {\em AAAI}.

\bibitem[\protect\citeauthoryear{Delbrouck and
  Dupont}{2017}]{Delbrouck:2017:arXiv}
Delbrouck, J.-B., and Dupont, S.
\newblock 2017.
\newblock Multimodal compact bilinear pooling for multimodal neural machine
  translation.
\newblock {\em arXiv}.

\bibitem[\protect\citeauthoryear{Devlin \bgroup et al\mbox.\egroup
  }{2019}]{bert2018}
Devlin, J.; Chang, M.; Lee, K.; and Toutanova, K.
\newblock 2019.
\newblock {BERT: Pre-training of Deep Bidirectional Transformers for Language
  Understanding}.
\newblock In {\em NAACL}.

\bibitem[\protect\citeauthoryear{Dou \bgroup et al\mbox.\egroup
  }{2018}]{Dou:2018:EMNLP}
Dou, Z.; Tu, Z.; Wang, X.; Shi, S.; and Zhang, T.
\newblock 2018.
\newblock Exploiting deep representations for neural machine translation.
\newblock In {\em EMNLP}.

\bibitem[\protect\citeauthoryear{Dou \bgroup et al\mbox.\egroup
  }{2019}]{Dou:2019:AAAI}
Dou, Z.; Tu, Z.; Wang, X.; Wang, L.; Shi, S.; and Zhang, T.
\newblock 2019.
\newblock Dynamic layer aggregation for neural machine translation with
  routing-by-agreement.
\newblock In {\em AAAI}.

\bibitem[\protect\citeauthoryear{Dozat and Manning}{2017}]{dozat2017ICLR}
Dozat, T., and Manning, C.~D.
\newblock 2017.
\newblock Deep biaffine attention for neural dependency parsing.
\newblock In {\em ICLR}.

\bibitem[\protect\citeauthoryear{Fukui \bgroup et al\mbox.\egroup
  }{2016}]{fukui2016multimodal}
Fukui, A.; Park, D.~H.; Yang, D.; Rohrbach, A.; Darrell, T.; and Rohrbach, M.
\newblock 2016.
\newblock {Multimodal Compact Bilinear Pooling for Visual Question Answering
  and Visual Grounding}.
\newblock In {\em EMNLP}.

\bibitem[\protect\citeauthoryear{Gao \bgroup et al\mbox.\egroup
  }{2016}]{Gao:2016:CVPR}
Gao, Y.; Beijbom, O.; Zhang, N.; and Darrell, T.
\newblock 2016.
\newblock Compact bilinear pooling.
\newblock In {\em CVPR}.

\bibitem[\protect\citeauthoryear{Goudreau \bgroup et al\mbox.\egroup
  }{1994}]{goudreau1994firstorder}
Goudreau, M.~W.; Giles, C.~L.; Chakradhar, S.~T.; and Chen, D.
\newblock 1994.
\newblock First-order versus second-order single-layer recurrent neural
  networks.
\newblock {\em IEEE Transactions on Neural Networks} 5(3):511--513.

\bibitem[\protect\citeauthoryear{He \bgroup et al\mbox.\egroup
  }{2016}]{he2016CVPR}
He, K.; Zhang, X.; Ren, S.; and Sun, J.
\newblock 2016.
\newblock Deep residual learning for image recognition.
\newblock In {\em CVPR}.

\bibitem[\protect\citeauthoryear{Kim \bgroup et al\mbox.\egroup
  }{2017}]{kim2016hadamard}
Kim, J.-H.; On, K.-W.; Lim, W.; Kim, J.; Ha, J.-W.; and Zhang, B.-T.
\newblock 2017.
\newblock Hadamard product for low-rank bilinear pooling.
\newblock In {\em ICLR}.

\bibitem[\protect\citeauthoryear{Koch, Poggio, and
  Torre}{1983}]{koch1983nonlinear}
Koch, C.; Poggio, T.; and Torre, V.
\newblock 1983.
\newblock Nonlinear interactions in a dendritic tree: localization, timing, and
  role in information processing.
\newblock {\em Proceedings of the National Academy of Sciences}
  80(9):2799--2802.

\bibitem[\protect\citeauthoryear{Koehn}{2004}]{Koehn2004Statistical}
Koehn, P.
\newblock 2004.
\newblock {Statistical Significance Tests for Machine Translation Evaluation}.
\newblock In {\em EMNLP}.

\bibitem[\protect\citeauthoryear{Kong and Fowlkes}{2017}]{Kong:2017:CVPR}
Kong, S., and Fowlkes, C.
\newblock 2017.
\newblock Low-rank bilinear pooling for fine-grained classification.
\newblock In {\em CVPR}.

\bibitem[\protect\citeauthoryear{Li \bgroup et al\mbox.\egroup
  }{2018}]{Li:2018:EMNLP}
Li, J.; Tu, Z.; Yang, B.; Lyu, M.~R.; and Zhang, T.
\newblock 2018.
\newblock {Multi-Head Attention with Disagreement Regularization}.
\newblock In {\em EMNLP}.

\bibitem[\protect\citeauthoryear{Li \bgroup et al\mbox.\egroup
  }{2019}]{Li:2019:NAACL}
Li, J.; Yang, B.; Dou, Z.-Y.; Wang, X.; Lyu, M.~R.; and Tu, Z.
\newblock 2019.
\newblock Information aggregation for multi-head attention with
  routing-by-agreement.
\newblock In {\em NAACL}.

\bibitem[\protect\citeauthoryear{Lin, RoyChowdhury, and
  Maji}{2015}]{Lin:2015:ICCV}
Lin, T.-Y.; RoyChowdhury, A.; and Maji, S.
\newblock 2015.
\newblock Bilinear cnn models for fine-grained visual recognition.
\newblock In {\em ICCV}.

\bibitem[\protect\citeauthoryear{Morcos and Harvey}{2016}]{Morcos:2016:Nature}
Morcos, A.~S., and Harvey, C.~D.
\newblock 2016.
\newblock History-dependent variability in population dynamics during evidence
  accumulation in cortex.
\newblock {\em Nature neuroscience} 19(12):1672.

\bibitem[\protect\citeauthoryear{Papineni \bgroup et al\mbox.\egroup
  }{2002}]{papineni2002bleu}
Papineni, K.; Roukos, S.; Ward, T.; and Zhu, W.-J.
\newblock 2002.
\newblock {BLEU: A method for Automatic Evaluation of Machine Translation}.
\newblock In {\em ACL}.

\bibitem[\protect\citeauthoryear{Peters \bgroup et al\mbox.\egroup
  }{2018}]{Peters:2018:NAACL}
Peters, M.~E.; Neumann, M.; Iyyer, M.; Gardner, M.; Clark, C.; Lee, K.; and
  Zettlemoyer, L.
\newblock 2018.
\newblock {{Deep Contextualized Word Representations}}.
\newblock In {\em NAACL}.

\bibitem[\protect\citeauthoryear{Pirsiavash, Ramanan, and
  Fowlkes}{2009}]{pirsiavash2009bilinear}
Pirsiavash, H.; Ramanan, D.; and Fowlkes, C.~C.
\newblock 2009.
\newblock Bilinear classifiers for visual recognition.
\newblock In {\em NIPS}.

\bibitem[\protect\citeauthoryear{Raganato and
  Tiedemann}{2018}]{raganato2018analysis}
Raganato, A., and Tiedemann, J.
\newblock 2018.
\newblock {An Analysis of Encoder Representations in Transformer-Based Machine
  Translation}.
\newblock In {\em EMNLP BlackboxNLP Workshop}.

\bibitem[\protect\citeauthoryear{Sennrich, Haddow, and
  Birch}{2016}]{sennrich2016neural}
Sennrich, R.; Haddow, B.; and Birch, A.
\newblock 2016.
\newblock {Neural Machine Translation of Rare Words with Subword Units}.
\newblock {\em ACL}.

\bibitem[\protect\citeauthoryear{Shen \bgroup et al\mbox.\egroup
  }{2018}]{Shen:2018:NAACL}
Shen, Y.; He, D.; Qin, T.; and Liu, T.-Y.
\newblock 2018.
\newblock Dense information flow for neural machine translation.
\newblock {\em NAACL}.

\bibitem[\protect\citeauthoryear{Shi, Padhi, and Knight}{2016}]{Shi:2016:EMNLP}
Shi, X.; Padhi, I.; and Knight, K.
\newblock 2016.
\newblock Does string-based neural mt learn source syntax?
\newblock In {\em EMNLP}.

\bibitem[\protect\citeauthoryear{Strubell \bgroup et al\mbox.\egroup
  }{2018}]{Strubell:2018:EMNLP}
Strubell, E.; Verga, P.; Andor, D.; Weiss, D.; and McCallum, A.
\newblock 2018.
\newblock {Linguistically-Informed Self-Attention for Semantic Role Labeling}.
\newblock In {\em EMNLP}.

\bibitem[\protect\citeauthoryear{Tenenbaum and
  Freeman}{2000}]{Tenenbaum:2000:NeuroComputation}
Tenenbaum, J.~B., and Freeman, W.~T.
\newblock 2000.
\newblock Separating style and content with bilinear models.
\newblock {\em Neural Computation} 12(6):1247--1283.

\bibitem[\protect\citeauthoryear{Tu \bgroup et al\mbox.\egroup
  }{2016}]{tu2016modeling}
Tu, Z.; Lu, Z.; Liu, Y.; Liu, X.; and Li, H.
\newblock 2016.
\newblock Modeling coverage for neural machine translation.
\newblock In {\em ACL 2016}.

\bibitem[\protect\citeauthoryear{Vaswani \bgroup et al\mbox.\egroup
  }{2017}]{Vaswani:2017:NIPS}
Vaswani, A.; Shazeer, N.; Parmar, N.; Uszkoreit, J.; Jones, L.; Gomez, A.~N.;
  Kaiser, {\L}.; and Polosukhin, I.
\newblock 2017.
\newblock {Attention Is All You Need}.
\newblock In {\em NIPS}.

\bibitem[\protect\citeauthoryear{Zhang \bgroup et al\mbox.\egroup
  }{2017}]{zhang2017thumt}
Zhang, J.; Ding, Y.; Shen, S.; Cheng, Y.; Sun, M.; Luan, H.; and Liu, Y.
\newblock 2017.
\newblock {THUMT: An Open Source Toolkit for Neural Machine Translation}.
\newblock {\em arXiv}.

\end{thebibliography}
\bibliographystyle{aaai}

\end{document}